\documentclass{article}
\usepackage{acra}

\usepackage[utf8]{inputenc}
\usepackage{amsmath, amssymb, amsthm}
\usepackage{graphicx}
\usepackage{hyperref}
\usepackage{cite}
\usepackage{booktabs}
\usepackage{xcolor}
\usepackage{tikz}
\usepackage{tabularx}
\usepackage{float}
\usepackage{caption} 
\usepackage{cleveref}

\title{Solomonoff-Inspired Hypothesis Ranking with LLMs for Prediction Under Uncertainty}

\author{
\scriptsize
\begin{minipage}[t]{0.24\textwidth}
    \centering
    \textbf{Josh Barber} \\
    Queensland University of Technology (QUT) \\
    Brisbane, Australia \\
    \texttt{josh.barber@connect.qut.edu.au}
\end{minipage}%
\hfill
\begin{minipage}[t]{0.24\textwidth}
    \centering
    \textbf{Rourke Young} \\
    Queensland University of Technology (QUT) \\
    Brisbane, Australia \\
    \texttt{rourke.young@connect.qut.edu.au}
\end{minipage}%
\hfill
\begin{minipage}[t]{0.24\textwidth}
    \centering
    \textbf{Cameron Coombe} \\
    Queensland University of Technology (QUT), CSIRO \\
    Brisbane, Australia \\
    \texttt{cameron.coombe@hdr.qut.edu.au}
\end{minipage}%
\hfill
\begin{minipage}[t]{0.24\textwidth}
    \centering
    \textbf{Will Browne} \\
    Queensland University of Technology (QUT) \\
    Brisbane, Australia \\
    \texttt{will.browne@qut.edu.au}
\end{minipage}%
}

\date{}

\begin{document}
\maketitle

\begin{abstract}
Reasoning under uncertainty is a key challenge in AI, especially for real-world tasks, where problems with sparse data demands systematic generalisation. Existing approaches struggle to balance accuracy and simplicity when evaluating multiple candidate solutions. We propose a Solomonoff-inspired method that weights LLM-generated hypotheses by simplicity and predictive fit. Applied to benchmark (Mini-ARC) tasks, our method produces Solomonoff-weighted mixtures for per-cell predictions, yielding conservative, uncertainty-aware outputs even when hypotheses are noisy or partially incorrect. Compared to Bayesian Model Averaging (BMA), Solomonoff scoring spreads probability more evenly across competing hypotheses, while BMA concentrates weight on the most likely but potentially flawed candidates. Across tasks, this highlights the value of algorithmic information-theoretic priors for interpretable, reliable multi-hypothesis reasoning under uncertainty.
\end{abstract}

\section{Introduction}

\noindent Humans rarely bet everything on a single guess. Whether crossing a busy street or interpreting vague instructions, we naturally keep multiple possibilities in mind before committing to an action. This ability to reason under uncertainty by simultaneously considering and weighing alternative outcomes is a hallmark of human cognition \cite{Herd2021, Friston2010FreeEnergy}. It allows us to make decisions that are robust even in unfamiliar or ambiguous situations, which will be a necessary component of Artificial Intelligence (AI) embedded in real-world robotics. \\

\noindent However, many modern AI systems struggle with uncertainty estimation. Large language models (LLMs) such as ChatGPT often produce outputs with high confidence \textit{even when incorrect}, due to miscalibrated uncertainty \cite{pelucchi2023chatgptpromptingestimatepredictive}. More generally, artificial neural networks frequently fail to provide reliable confidence estimates, leading to over- or under-confident predictions in complex domains, i.e., domains that expose multiple hypotheses \cite{Gawlikowski2023}. These limitations motivate the development of methods that explicitly evaluate and integrate multiple hypotheses. An ideal testbed for exploring such mechanisms needs to be open-ended, contain multiple hypotheses, and not be exhaustive by design. The Abstraction and Reasoning Corpus (ARC) \cite{chollet2019measure}, also referred to as ARC-AGI-1, is a collection of grid-based visual reasoning tasks in which an agent must infer the transformation between an input and corresponding output grids, designed to evaluate systematic generalisation rather than pattern memorisation. Mini-ARC \cite{kim2022playgrounds}, a simplified version of ARC, reduces problem complexity while retaining the core challenge: given several input–output examples, an agent must infer the hidden transformation and apply it to a novel input. An example Mini-ARC task is shown in Figure~\ref{fig:1}, where objects stretch upward in even columns and downward in odd columns. Mini-ARC thus provides a controlled environment to investigate multi-hypothesis reasoning and predictive uncertainty. \\

\noindent Uncertainty in AI has been addressed with probabilistic and Bayesian models \cite{gal2016dropout} and neural network methods like ensembles and Monte Carlo dropout \cite{lakshminarayanan2017simple, malinin2018predictive}, though these struggle in complex, combinatorial domains common to robotics. Tasks such as motion planning or perception often involve multiple plausible outcomes that must be weighed before action. In ARC and Mini-ARC, similar uncertainty arises when evaluating competing transformations \cite{chollet2019measure, kim2022playgrounds, singhal2024conceptsearch}, with some systems focusing on a single hypothesis while others explore multiple alternatives \cite{tan2023llm_multiple_agents_arc}. Although multi-hypothesis reasoning can improve generalisation \cite{choi2023hypothesis}, integrating and weighting competing hypotheses remains challenging. Frameworks combining principled hypothesis evaluation with interpretable scoring, such as the Solomonoff-inspired approach proposed here, could extend to domains requiring multi-hypothesis reasoning, such as path planning under uncertainty \cite{wang2023multiplehypothesispathplanninguncertain}. \\

\noindent Instead of selecting one hypothesis, we use a Solomonoff-inspired approach (see Section \ref{sec:solomonoffinduction}) that builds a probability distribution over candidates, weighting them by simplicity and consistency. This allows reasoning under ambiguity, integrating alternatives, and making more robust predictions than single-hypothesis methods. \\

\noindent The contributions of this work are:
\begin{itemize}
    \item \textbf{Solomonoff-inspired scoring:} Assigns probabilities to LLM-generated hypotheses using \emph{simplicity} and \emph{predictive accuracy}, unlike prior single-hypothesis methods, enabling robust evaluation of alternatives.
    
    \item \textbf{Weighted mixture-based prediction:} Aggregates multiple hypotheses in a \emph{Solomonoff-weighted matrix} with cell-level uncertainty, allowing informed predictions under ambiguity.
    
\item \textbf{Empirical insights:} Compared to Bayesian Model Averaging (BMA), the Solomonoff method was better calibrated in its confidence (uncertain when appropriate), and showed less noise across hypotheses.

\end{itemize}

\noindent The words ``problem,” ``task,” and ``puzzle” all refer to a single Mini-ARC puzzle consisting of several input-output grid examples and a test pair. These terms are used inter-changeably.

\section{Related Work}

\noindent Bayesian methods are widely used to model predictive uncertainty in deep learning \cite{Jospin2022, gal2016dropout, Blundell2015}. By placing distributions over parameters, Bayesian neural networks capture both model uncertainty (epistemic), arising from limited data or incomplete knowledge, and data uncertainty (aleatoric), arising from inherent noise in observations. Extensions such as ensembles and Monte Carlo dropout apply these principles to deep architectures \cite{lakshminarayanan2017simple, malinin2018predictive}. However, Bayesian methods are computationally expensive, sensitive to priors, and less effective for high-dimensional, combinatorial inputs \cite{Kendall2017, Wenzel2020}. In ARC-style tasks, the main challenge is structural uncertainty (uncertainty over transformations or object manipulations) rather than parameter uncertainty, which standard Bayesian networks handle at the parameter level. Structural uncertainty instead arises from not knowing the rules or operations generating the observed patterns. \\

\noindent Multi-hypothesis and LLM-based approaches target this structural uncertainty more directly. Wang et al.~\cite{wang2024hypothesissearchinductivereasoning} show that maintaining multiple candidate hypotheses, ranked by simplicity and data consistency, improves robustness and generalisation compared to committing early to a single explanation. This highlights the value of reasoning over multiple alternatives, especially with sparse training examples as in Mini-ARC. \\

\noindent Our method evaluates LLM-generated hypotheses with a Solomonoff-inspired scoring scheme, weighting them by simplicity and predictive accuracy. As a probabilistic baseline, we compare against Bayesian Model Averaging (BMA)~\cite{hoeting1999bayesian,DUAN20071371}, which combines competing models probabilistically, weighting better-performing predictions higher. As a well-established method, it provides a reliable benchmark. Unlike parameter-based Bayesian methods, our approach targets structural uncertainty by ranking and merging alternative hypotheses, making it suitable for combinatorial reasoning with limited data. \\

\subsection{Problem Setting}

\begin{figure}[H]
    \centering
    \includegraphics[width=0.85\linewidth]{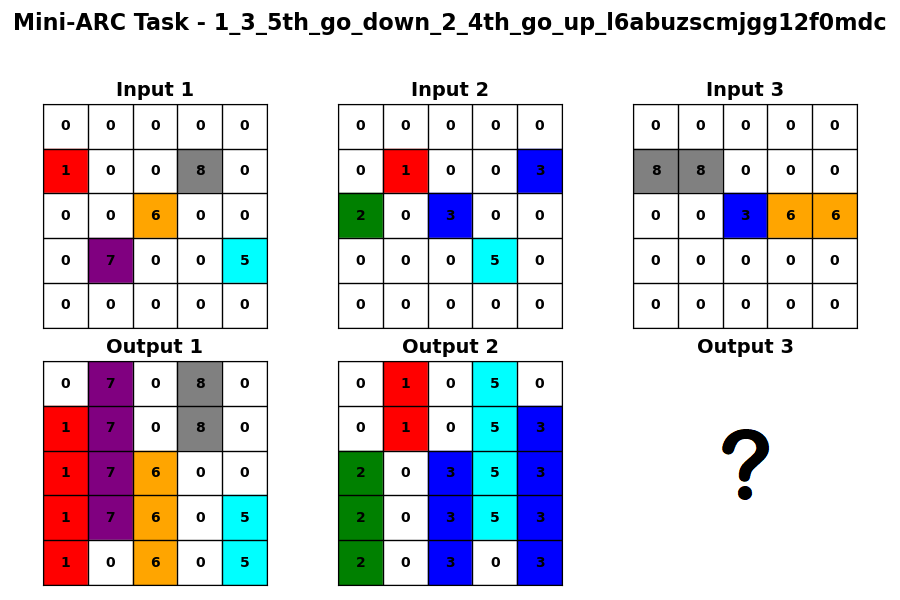}
    \caption{Mini-ARC task where odd columns shift down and even columns up; the agent infers this rule from input–output examples (see Table~\ref{tab:task-a-hypotheses}). The nth output is withheld for illustration but included in the dataset.}
    \label{fig:1}
\end{figure}

\noindent Mini-ARC \cite{kim2022playgrounds} is a simplified version of the \textit{Abstraction and Reasoning Corpus} (ARC) \cite{chollet2019measure}, designed to test abstract reasoning and systematic generalisation in small, grid-based visual tasks. Each problem consists of \textit{input–output pairs}, where the input is a small grid (typically $5 \times 5$) of integer-valued colours (0–9, with 0 as background), and the output shows the result of a hidden transformation. The agent must infer the transformation and predict outputs for novel inputs (see Figure \ref{fig:1}). \\

\noindent Compared to ARC, Mini-ARC reduces combinatorial complexity with smaller grids (up to 30x30 vs 5x5), fewer input–output pairs (2-10 vs 3-5), and minimalistic settings, making tasks more tractable, interpretable, and suitable for evaluating abstract reasoning, systematic generalisation, and multi-hypothesis inference \cite{kim2022playgrounds}.

\noindent Mini-ARC tasks emphasise object-centric reasoning, where agents detect discrete objects (connected regions of the same colour) and track changes in properties such as size, position, and colour. However, reasoning is not restricted to objects alone: transformations may also involve grid-level or relational operations, including translation, reflection, rotation, recolouring, or compositional rules (e.g., objects stretching in opposite vertical directions depending on column order, Figure~\ref{fig:1}). With few examples per task, agents must rely on abstract reasoning and weigh multiple plausible hypotheses rather than memorising outputs. \\

\noindent Mini-ARC provides a complex and structured environment to study hypothesis-driven reasoning under epistemic uncertainty. Generating and evaluating multiple candidate hypotheses allows systems to handle uncertainty arising from incomplete knowledge, compare competing explanations, and integrate them into robust predictions for novel inputs, making Mini-ARC an ideal benchmark for testing uncertainty-aware reasoning \cite{kim2022playgrounds}.

\subsection{Solomonoff Induction}
\label{sec:solomonoffinduction}

\noindent Unlike standard predictive methods, Solomonoff induction offers a universal, theoretically optimal framework for reasoning under uncertainty \cite{Solomonoft1975InductiveInference}. It considers all (infinite) possible computer programs that could generate the observed data and weights them according to simplicity, assigning higher prior likelihood to shorter, more concise explanations. This embodies \emph{Occam's razor} \cite{ockham1323}, the principle that simpler hypotheses should be preferred when multiple explanations are possible. \\

\noindent Formally, for a finite binary string $x$, the \emph{Solomonoff prior} assigns probability
\begin{equation}
P(x) = \sum_{p: U(p)=x*} 2^{-|p|},
\end{equation}
where the sum is over all programs $p$ on a universal prefix-free Turing machine $U$ that output a string starting with $x$, and $|p|$ is the length of $p$ in bits. Programs with fewer bits contribute more to the sum, reflecting the principle that simpler explanations are preferred. \\

\noindent From this prior, the probability of the next symbol $a$ given the observed string $x$ is
\begin{equation}
P(a \mid x) = \frac{P(xa)}{P(x)}.
\end{equation}

\noindent Solomonoff induction assigns higher weight to hypotheses that are both simple and consistent with observed data. It formalises the balance between \textit{fit to the data} and \textit{simplicity of explanation}, providing a principled framework for inductive inference. In practice, however, the method is incomputable, since it requires summing over an infinite set of hypotheses (all possible programs).

\section{Method}

\noindent Solomonoff induction is theoretically optimal over infinite programs but incomputable in practice. We approximate it by using a finite set of LLM-generated, object-centered hypotheses for Mini-ARC tasks. Each hypothesis is scored for \textbf{simplicity} (token length) and \textbf{predictive accuracy} (cell-wise correctness) on training examples. Scores are then normalised into probabilities to form a \textbf{Solomonoff-weighted matrix}, giving per-cell likelihoods.

\subsection{Hypotheses Generation}

\noindent Candidate hypotheses describe possible transformations from input to output for a given ARC problem. 
We first extract object-level features from the input and output grids using a domain specific language (DSL) \cite{hodel2023arc}, identifying contiguous objects and their attributes such as position, shape, and colour composition. These objects are then serialised into structured representations, including spatial relations between objects, which serve as input to a LLM. The LLM generates hierarchical hypotheses, consisting of an overall transformation description as well as sub-hypotheses for individual objects, capturing both object-level changes and higher-level grid patterns. \\

\noindent For example, consider a Mini-ARC task where the input grid contains a single red square in the top-left corner, and the output grid shows the same square translated two cells to the right. An LLM-generated hypothesis could describe this transformation as:

\begin{itemize}
\item \textbf{Overall transformation:} ``Shift all red squares two cells to the right."
\item \textbf{Sub-hypothesis for the red square:} ``Move from position (0,0) to (0,2) without changing size or colour."
\end{itemize} 

\noindent Another plausible hypothesis for the same task could be:

\begin{itemize}
\item \textbf{Overall transformation:} ``Copy each object and shift it horizontally by two cells."
\item \textbf{Sub-hypothesis for the red square:} ``Duplicate the red square at a horizontal offset of +2, keeping the original in place."
\end{itemize}

\noindent We employ GPT-4.0 for hypothesis generation due to its enhanced abstract reasoning capabilities over GPT-3.5, as demonstrated by superior performance in zero-shot reasoning benchmarks \cite{arxiv2305.12477}. While the LLM generates candidate hypotheses from the training data, it does not provide explicit scores or likelihoods. These hypotheses are evaluated and ranked using a Solomonoff-inspired prior combined with a data-fit metric. For this work, object-based transformations were prioritised to test our weighting scheme; while not optimal for ARC in general, this provided a controlled setting to assess the approach.

\subsection{Per-hypothesis Solomonoff-inspired score}

\noindent We develop a per-hypothesis scoring scheme that is explicitly designed to be Solomonoff-inspired while remaining computationally tractable for a finite set of hypotheses. For a single hypothesis $h$ (a human/LLM readable description plus sub-steps in the set of possible hypotheses $\mathcal H$) we compute three quantities: a description length proxy $L(h)$, a normalised simplicity score $\mathrm{Simplicity}(h)$, and a data likelihood proxy $\mathrm{Accuracy}(h)$. Here, $h$ denotes the hypothesis being scored, while $h'$ is a dummy variable ranging over all hypotheses in $\mathcal H$ when computing global quantities such as minimum or maximum length. These are combined multiplicatively into a (normalised) Solomonoff-inspired score $\mathcal{S}(h)$. \\

\subsubsection{Description length and simplicity} 

The classical Solomonoff method uses the exact program length $|p|$ to set the prior weight $2^{-|p|}$. We replace exact program length with a practical proxy: the token length $L(h)$ of the hypothesis text. Let
\begin{equation}
L_{\min} = \min_{h' \in \mathcal H} L(h'), \qquad
L_{\max} = \max_{h' \in \mathcal H} L(h').
\end{equation}

\noindent Recent work has introduced the notion of \emph{token complexity} \cite{xu2025tokencomplexity}, which characterises the minimal number of tokens required for successful problem-solving. This concept provides a principled way to reason about the efficiency of generated reasoning chains and supports the idea that shorter, simpler hypotheses can be preferable, provided they retain sufficient information for accurate prediction. \\ 

\noindent We therefore map token length (computed from the hypothesis and its sub-hypotheses using GPT-style tokenisation via the \texttt{tiktoken} library) into a normalised \emph{simplicity score} between $[0,1]$ using a clipped linear transform:
\begin{equation}
\label{eq:simplicity}
\mathrm{Simplicity}(h)
=
1 - \frac{L(h) - L_{\min}}{L_{\max} - L_{\min}}.
\end{equation}

\subsubsection{Data-fit (likelihood) proxy}  

\noindent To measure how well a hypothesis $h$ explains the training data, we compute its \emph{cell-wise accuracy} over all cells in the output grids. That is, we evaluate every cell, including background cells, to determine the fraction of positions where the predicted and true values match. \\

\noindent Let the full dataset contain $n$ examples. Using leave-one-out cross-validation, we define the training set as the first $n-1$ examples and the evaluation set as the $n$-th (held-out) example. For example $i$ in the training set, let $Y^{(i)}$ denote the expected output grid and $\hat Y^{(i)}_h$ the grid predicted by hypothesis $h$. Each grid is a two-dimensional array of cells, where $r$ indexes the rows and $c$ indexes the columns. \\

\noindent The accuracy of hypothesis $h$ is then the proportion of correctly predicted cells across all training examples. This gives a value in $[0,1]$ and directly reflects how well $h$ reproduces the observed data, with higher values assigned to hypotheses that better match the complete output grids. Unlike likelihoods based on specific noise models, this measure is fully nonparametric and purely empirical. \\

\subsubsection{Per-hypothesis Solomonoff-inspired score}  

Combine simplicity and accuracy multiplicatively to obtain the score used:
\begin{equation}
\mathcal{S}(h)\;=\;\mathrm{Simplicity}(h)\times\mathrm{Accuracy}(h)
\label{eq:solomonoff_score}
\end{equation}

\noindent This product serves as an normalised posterior-like quantity: $\mathrm{Simplicity}(h)$ plays the role of a prior and $\mathrm{Accuracy}(h)$ plays the role of a likelihood. \\ 

\noindent If a proper probabilistic interpretation is needed, one can normalise over the generated hypothesis set:
\begin{equation}
\tilde P(h\mid D)\;=\;\frac{\mathcal{S}(h)}{\sum_{h'\in\mathcal H} \mathcal{S}(h')},
\end{equation}
yielding a discrete distribution over the finite hypothesis pool. \\

\subsubsection{Finite, LLM-based Solomonoff-inspired Scoring}

\noindent Instead of the classical Solomonoff mixture that sums over \emph{all} computable programs with prior $2^{-|p|}$, we (1) restrict to a finite, data-conditioned hypothesis set $\mathcal H$ produced by an LLM, (2) replace the exact universal prior with a practical simplicity proxy derived from token length, and (3) replace an intractable universal mixture with per-hypothesis scoring. \\

\noindent Despite these simplifications, the scheme preserves the two core Solomonoff principles:

\begin{enumerate}
  \item \textbf{Occam bias (simplicity prior):} shorter hypotheses are given higher prior weight via $\mathrm{Simplicity}(h)$, a monotone mapping of description length.  
  \item \textbf{Data-driven posterior (likelihood):} hypotheses are preferred if they explain the observed data well via $\mathrm{Accuracy}(h)$.  
\end{enumerate}

\noindent Multiplying these two terms yields a normalised posterior-like score that favours simple hypotheses that fit the data, exactly the behaviour Solomonoff formalises, but in a form that is computable on a finite hypothesis set. \\

\noindent Operationally, this has several consequences: 

\begin{itemize}
    \item \textbf{Tractability:} we never need to enumerate or sum over the (infinite) set of all programs, unlike traditional Solomonoff induction approaches. Scoring is only done for hypotheses actually generated by the LLM.  
    \item \textbf{Transparency and debugging:} per-hypothesis scores decompose into simplicity and accuracy, facilitating analysis of failure modes.  
    \item \textbf{Dependence on hypothesis generation:} the quality of the approximation depends on $\mathcal H$; if the LLM fails to propose a correct class of programs, no scoring scheme can recover it.  
    \item \textbf{Loss of universality:} unlike Solomonoff's universal predictor \cite{Solomonoft1975InductiveInference,Hutter:24uaibook2}, this method is not universal. Its guarantees depend on the hypothesis generator and tokeniser/length proxy. Some true transformations may be missing from the generated set, and even plausible hypotheses can be overlooked, as shown in the Results and Discussion sections.
\end{itemize}

\subsection{Solomonoff-weighted Matrix}

\subsubsection{Theory}

\noindent Given a set of hypotheses $\mathcal{H}$ with per-hypothesis scores $\mathcal{S}(h)$, we construct a \textit{Solomonoff-weighted matrix} for a given input Mini-ARC grid to capture cell-level predictive uncertainty. Let the input grid be of size $R \times C$ (rows $\times$ columns) and let each hypothesis $h$ predict a grid $\hat Y_h$ of the same size. \\

\noindent Define the per-cell probability of value $v$ at position $(r,c)$ as the weighted sum over hypotheses:
\begin{equation}
\label{7}
P_{r,c}(v) = \sum_{h \in \mathcal{H}} \tilde P(h\mid D) \, \mathbf{1}\{\hat Y_h(r,c) = v\},
\end{equation}
where $\tilde P(h \mid D)$ is the normalised per-hypothesis Solomonoff-inspired score
\begin{equation}
\label{8}
\tilde P(h\mid D) = \frac{\mathcal{S}(h)}{\sum_{h'\in \mathcal{H}} \mathcal{S}(h')},
\end{equation}
and $\mathbf{1}\{\cdot\}$ is the indicator function. \\

\noindent Intuitively, $P_{r,c}(v)$ represents the probability that cell $(r,c)$ takes value $v$ under the mixture of hypotheses, giving more weight to simpler and more accurate hypotheses. The resulting matrix
\begin{equation}
\mathbf{P} = \big[P_{r,c}(v)\big]_{r=1..R,\, c=1..C,\, v \in \mathcal{V}}
\end{equation}
where $\mathcal{V}$ is the set of possible cell values, encodes a full distribution over outputs for the input grid. \\

\subsubsection{Prediction using the weighted matrix} 

\noindent The final predicted grid can be obtained by choosing the most likely value at each cell:
\begin{equation}
\hat Y_{r,c} = \arg\max_{v \in \mathcal{V}} P_{r,c}(v).
\end{equation}

\noindent This approach preserves uncertainty: if multiple hypotheses disagree on a cell, the distribution $P_{r,c}$ reflects this, instead of committing prematurely to a single hypothesis. The Solomonoff-weighted matrix thus transforms per-hypothesis scores into robust, weighted-mixture predictions that capture uncertainty, making the method suitable for Mini-ARC tasks.

\section{Experimental Setup}

\noindent In our analysis, we employ Bayesian Model Averaging (BMA) as a comparative framework. Whereas Solomonoff scoring prioritises hypotheses according to algorithmic simplicity and an associated universal prior, BMA instead weights hypotheses according to their ability to explain the training data. Each hypothesis $h_i$ is assigned a posterior weight  
\begin{equation}
    w_i \;=\; \frac{P(D \mid h_i) \, P(h_i)}{\sum_j P(D \mid h_j) \, P(h_j)},
\end{equation}
\noindent where $P(D \mid h_i)$ is the likelihood of the training data $D$ under hypothesis $h_i$, and $P(h_i)$ is the prior over hypotheses (taken to be uniform in our experiments). The weights are normalised across all hypotheses so that $\sum_i w_i = 1$. \\ 

\noindent To compute the likelihood, we model each grid cell as a categorical observation with error probability $\epsilon$. Given a predicted output grid $\hat{y}$ from hypothesis $h_i$ and the ground-truth grid $y$, the per-cell likelihood is \\ 
\begin{equation}
P(y_{rc} \mid h_i) \;=\;
\begin{cases}
1-\epsilon, & y_{rc} = \hat{y}_{rc}, \\
\frac{\epsilon}{K-1}, & y_{rc} \neq \hat{y}_{rc},
\end{cases}
\end{equation}
\noindent where $K$ is the number of possible colours. The overall likelihood is then \\ 
\begin{equation}
P(D \mid h_i) \;=\; (1-\epsilon)^{N_{\text{correct}}} \left(\frac{\epsilon}{K-1}\right)^{N_{\text{wrong}}},
\end{equation}
\noindent with $N_{\text{correct}}$ and $N_{\text{wrong}}$ denoting the number of correctly and incorrectly predicted cells, respectively.  \\

\noindent This produces weighted prediction matrices capturing the model’s per-cell confidence. Comparing BMA- and Solomonoff-weighted predictions benchmarks a practical, data-driven approach against a simplicity-based prior, revealing where empirical evidence aligns or diverges from theoretical rankings. Using the same hypothesis set for both methods allows direct comparison of their ranking strategies and handling of uncertainty. \\

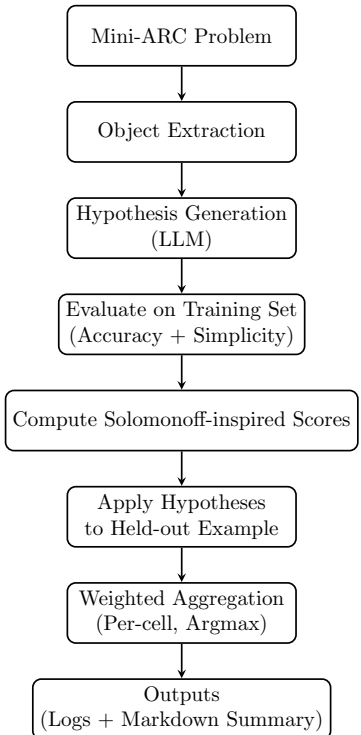
\begin{figure}[H]
    \centering
    \scalebox{0.8}{ 
    \begin{tikzpicture}[node distance=1.6cm, >=stealth, thick]
        \tikzstyle{block} = [draw, rounded corners, align=center, minimum width=3.8cm, minimum height=1cm]
        \tikzstyle{arrow} = [->, thick]

        \node (input) [block] {Mini-ARC Problem};
        \node (objects) [block, below of=input] {Object Extraction};
        \node (hypgen) [block, below of=objects] {Hypothesis Generation \\ (LLM)};
        \node (eval) [block, below of=hypgen] {Evaluate on Training Set \\ (Accuracy + Simplicity)};
        \node (score) [block, below of=eval] {Compute Solomonoff-inspired Scores};
        \node (apply) [block, below of=score] {Apply Hypotheses \\ to Held-out Example};
        \node (aggregate) [block, below of=apply] {Weighted Aggregation \\ (Per-cell, Argmax)};
        \node (output) [block, below of=aggregate] {Outputs \\ (Logs + Markdown Summary)};

        \draw[arrow] (input) -- (objects);
        \draw[arrow] (objects) -- (hypgen);
        \draw[arrow] (hypgen) -- (eval);
        \draw[arrow] (eval) -- (score);
        \draw[arrow] (score) -- (apply);
        \draw[arrow] (apply) -- (aggregate);
        \draw[arrow] (aggregate) -- (output);
    \end{tikzpicture}
    } 
    \caption{Overall experimental pipeline: (1) Input task and object extraction, (2) Hypothesis generation via LLM, (3) Evaluation using accuracy and simplicity, (4) Solomonoff-inspired or Bayesian weighting, and (5) Aggregated per-cell predictions.}
    \label{fig:experiment-pipeline}
\end{figure}

\noindent In the Solomonoff approach we evaluate how well Solomonoff-inspired scoring can rank and combine hypotheses generated by an LLM, using the GPT API to issue GPT-4 prompts, for Mini-ARC tasks, where Figure~\ref{fig:experiment-pipeline} illustrates the overall pipeline. In the BMA method, we replace Solomonoff-based scoring with BMA-based scoring. \\

\noindent The procedure consists of four stages: \ 

\noindent \textbf{Input and object extraction.} Candidate hypotheses describe possible transformations for an ARC problem. Objects are extracted and serialised with attributes and spatial relations, then fed to a LLM. The LLM generates hierarchical hypotheses with an overall transformation plus sub-hypotheses for individual objects, capturing both object-level and grid-level patterns.  \

\noindent \textbf{Hypothesis generation.} GPT-4 uses structured object descriptions from the training set to propose $n=6$ candidate hypotheses, balancing diversity and evaluation cost. 6 hypotheses were chosen to test multiple hypotheses without occurring additional time/financial costs. Each hypothesis contains a concise natural language description of the global transformation and sub-hypotheses for individual objects, capturing local or relational patterns. Outputs are requested in JSON format for deterministic parsing.  \

\noindent \textbf{Hypothesis evaluation and scoring.} Each hypothesis $h$ is applied back to the training examples to produce $\hat Y_i^h$. We compute accuracy (fraction of non-background cells correctly predicted), simplicity (inverse function of token length, Eq.~\eqref{eq:simplicity}), and the Solomonoff-inspired score $\mathcal{S}(h)$ (Eq.~\eqref{eq:solomonoff_score}). This yields a ranked list of hypotheses with per-example statistics.  \

\noindent \textbf{Prediction on held-out examples.} To test generalisation, we perform leave-one-out evaluation. Each held-out input $X_j$ is transformed by all $h \in \mathcal{H}$ to produce $\hat Y_j^h$, which are aggregated into a Solomonoff-weighted matrix via Eq. \eqref{7} and \eqref{8}.

\noindent The final prediction $\hat Y_j$ is obtained via per-cell argmax. Further, all data is logged and cached, (e.g. GPT calls).

\section{Results}

\begin{figure}[H]
    \centering
    \includegraphics[width=0.9\linewidth]{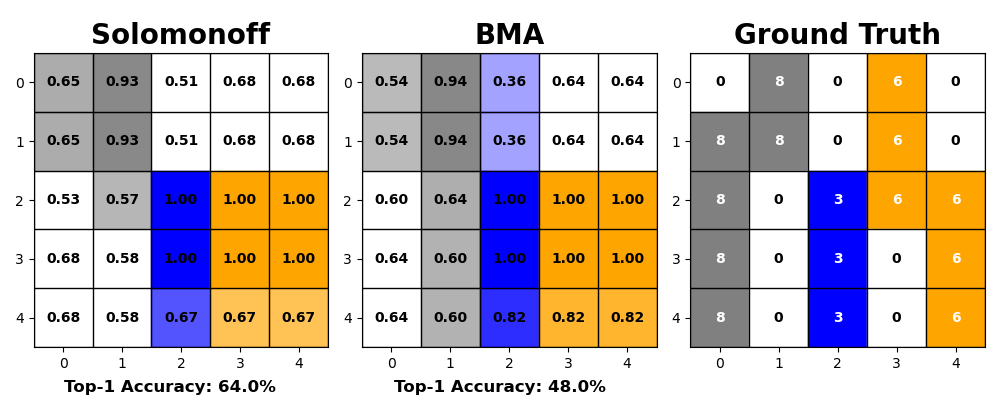}
    \caption{Task A - Alternating column shifts with six generated hypotheses. Comparison of Solomonoff-weighted vs Bayesian Model Averaging (BMA) predictions. Solomonoff achieved 64\% accuracy vs 48\% for BMA.}
    \label{fig:task-a-weighted-matrix}
\end{figure}

\begin{figure}[H]
    \centering
    \includegraphics[width=1\linewidth]{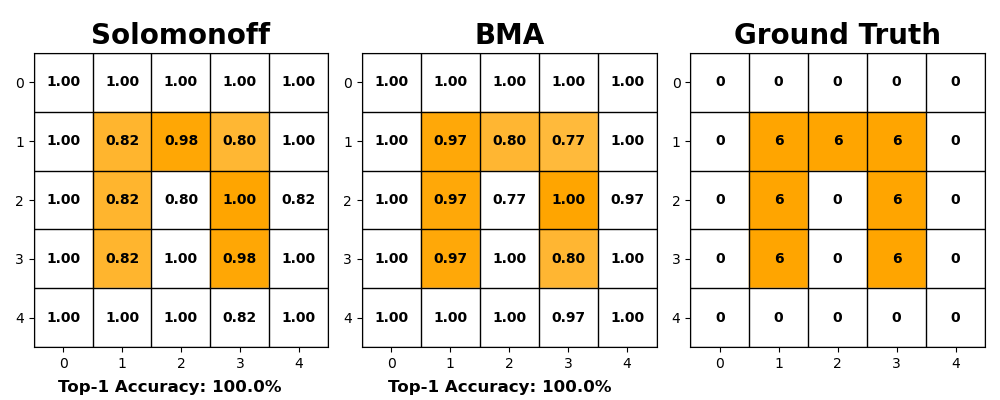}
    \caption{Task B - Centralisation task. Both Solomonoff and BMA predictions achieve 100\% accuracy, though Solomonoff shows more conservative confidence calibration.}
    \label{fig:task-b-weighted-matrix}
\end{figure}

\begin{figure}[H]
    \centering
    \includegraphics[width=1\linewidth]{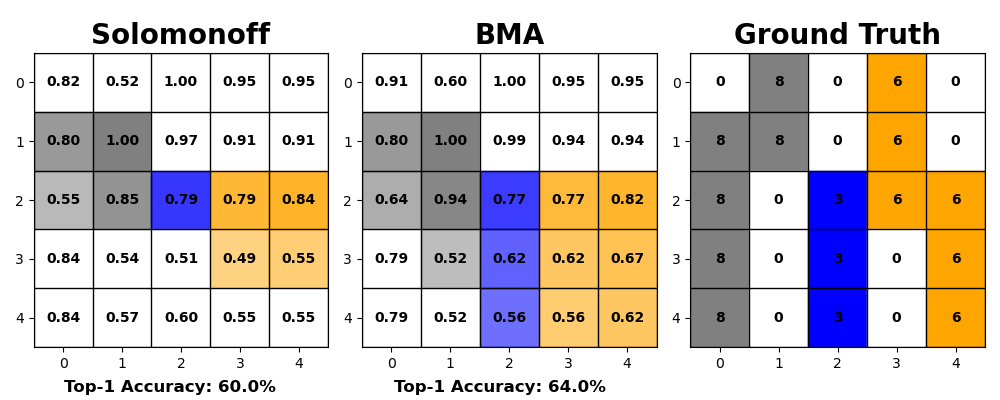}
    \caption{Task C – Alternating column shifts with 20 noisy hypotheses. Accuracy drops for both methods (Solomonoff 60\%, BMA 64\%), highlighting overconfidence versus cautious weighting under hypothesis noise. Compared to Task A, overall confidence and correctness are lower.}

    \label{fig:task-c-weighted-matrix}
\end{figure}

Figures~\ref{fig:task-a-weighted-matrix}--\ref{fig:task-c-weighted-matrix} show three Mini-ARC tasks as $5 \times 5$ grids. Solomonoff-inspired and BMA predictions are compared in the first two panels, with cell colour, transparency, and numbers showing label, confidence, and probability. The third panel gives ground truth with top-1 accuracy below. Note that Solomonoff and BMA differ in how sharply probabilities concentrate, especially under noisy hypotheses. Tables~\ref{tab:task-a-hypotheses}--\ref{tab:task-c-hypotheses} list ranked hypotheses, highlighting how weights distribute across candidates and shape predictive performance.

\begin{table}[H]
\centering
\tiny
\setlength{\tabcolsep}{2pt} 
\caption{Task A - Ranked Hypotheses}
\begin{tabular}{c|p{3.2cm}|c|c} 
\toprule
\textbf{Number} & \textbf{Hypothesis} & \textbf{Solomonoff Weight} & \textbf{BMA Weight} \\
\midrule
1 & Different cells in the input grid replicate vertically while maintaining column locations. & 0.245 & 0.300 \\
2 & Objects propagate vertically up and down until blocked. & 0.179 & 0.135 \\
3 & Objects duplicated vertically across grid height. & 0.178 & 0.300 \\
4 & Objects expand vertically to top/bottom while preserving column. & 0.173 & 0.165 \\
5 & Objects duplicated from top to bottom across grid. & 0.152 & 0.041 \\
6 & Objects translated vertically, repeating in straight lines. & 0.073 & 0.061 \\
\bottomrule
\end{tabular}
\label{tab:task-a-hypotheses}
\end{table}

\begin{table}[H]
\centering
\tiny
\setlength{\tabcolsep}{2pt} 
\caption{Task B - Ranked Hypotheses}

\begin{tabular}{c|p{3.2cm}|c|c} 
\toprule
\textbf{Rank} & \textbf{Hypothesis} & \textbf{Solomonoff Weight} & \textbf{BMA Weight} \\
\midrule
1 & Objects moved from original location to grid center, maintaining pattern. & 0.265 & 0.198 \\
2 & Cross-shaped clusters repositioned to grid center. & 0.215 & 0.133 \\
3 & Unique shape rotates 90° counterclockwise. & 0.183 & 0.033 \\
4 & Objects move toward center, shape and color preserved. & 0.181 & 0.242 \\
5 & Object shifts diagonally top-left to center. & 0.135 & 0.198 \\
6 & Objects repositioned into central '+' shape. & 0.021 & 0.198 \\
\bottomrule
\end{tabular}
\label{tab:task-b-hypotheses}
\end{table}

\begin{table}[H]
\centering
\tiny
\setlength{\tabcolsep}{2pt} 
\caption{Task C - Ranked Hypotheses (Abbreviated)}
\begin{tabular}{c|p{3.2cm}|c|c} 
\toprule
\textbf{Rank} & \textbf{Hypothesis} & \textbf{Solomonoff Weight} & \textbf{BMA Weight} \\
\midrule
1  & Objects duplicated downward until blocked. & 0.081 & 0.098 \\
2  & Objects replicated vertically in output grid. & 0.081 & 0.024 \\
3  & Downward replication based on object color counts. & 0.076 & 0.080 \\
4  & Objects fill columns top to bottom. & 0.074 & 0.044 \\
5  & Column filled with object copies, same color/position. & 0.072 & 0.080 \\
6  & Each object extended vertically across grid height. & 0.067 & 0.066 \\
7  & Colored objects expand upward until blocked. & 0.062 & 0.036 \\
8  & Objects duplicated above and below until filled. & 0.061 & 0.013 \\
9  & Objects repeated in column across all rows. & 0.059 & 0.054 \\
10 & Objects extend vertically to cover column. & 0.058 & 0.044 \\
11 & Objects expand up/down by 2–4 cells depending on relations. & 0.050 & 0.080 \\
12 & Objects expand downward, or upward if at bottom. & 0.047 & 0.080 \\
13 & Objects shifted/expanded vertically based on relations. & 0.047 & 0.044 \\
14 & Vertical replications depend on object position. & 0.045 & 0.080 \\
15 & Objects arranged into columns by type. & 0.043 & 0.011 \\
16 & Vertical replication equals initial row index + 1. & 0.042 & 0.020 \\
17 & Objects replicated horizontally to row edges. & 0.038 & 0.007 \\
18 & Colored cells expand vertically until blocked. & 0.036 & 0.036 \\
19 & Distinct colors duplicated in vertical columns. & 0.034 & 0.036 \\
20 & Objects stretch downward; overlaps overwrite. & 0.008 & 0.080 \\
\bottomrule
\end{tabular}
\label{tab:task-c-hypotheses}
\end{table}

\section{Discussion}

\noindent While the ranking results reveal differences between the Solomonoff and BMA methods, the generated hypotheses themselves are imperfect. Many candidates failed to capture the full structure of Mini-ARC tasks, containing redundancies, overly specific formulations, or partial errors. For instance, Table~\ref{tab:task-c-hypotheses} shows multiple hypotheses describing vertical replication with minor variations, some of which are inaccurate (see Tables \ref{tab:task-a-hypotheses}-\ref{tab:task-c-hypotheses}). An ideal hypothesis, in contrast, would fully and correctly capture the underlying transformation in a concise manner. Such a hypothesis would receive a higher Solomonoff score because it balances accuracy, reproducing the observed input-output patterns correctly, and simplicity, providing the minimal description necessary to explain the transformation. This highlights the inherent noise in LLM-generated hypotheses. No single candidate perfectly describes the underlying transformation, and producing higher-quality candidates typically requires extensive prompt engineering. \\

\noindent The key factor, therefore, is not the quality of individual hypotheses but how probability is distributed across the set. Both Solomonoff and BMA methods convert noisy, imperfect candidates into weighted mixtures, enabling robust predictions even when no single hypothesis is fully correct, with performance ultimately shaped by the quality of the generated candidate set. \\

\subsection{Comparison of Hypotheses Ranking}

The Solomonoff-inspired method and BMA differ in ranking hypotheses. Solomonoff scoring rewards simplicity: shorter, general descriptions get higher weight, even with similar accuracy. In Table~\ref{tab:task-b-hypotheses}, Solomonoff ranks ``Objects moved from original location to grid centre, maintaining pattern'' first (0.265), while BMA favours the longer ``Objects move toward centre, shape and colour preserved'' (0.242). Accuracy is similar (0.88 vs 0.90), but Solomonoff prefers brevity. BMA weights mainly by likelihood, often producing ties across equally accurate hypotheses, which Solomonoff avoids. \\

\noindent Table~\ref{tab:task-a-hypotheses} shows a similar trend: Solomonoff prefers ``Different cells replicate vertically while maintaining column locations'' (0.245), while BMA splits weight between it and ``Objects duplicated vertically across grid height'' ($\approx$0.30 each). Under noise (Table~\ref{tab:task-c-hypotheses}), Solomonoff concentrates weight on a few compact hypotheses (0.081 each), whereas BMA spreads mass across many variants. \\

\noindent Overall, Solomonoff stabilises predictions with concise rules, while BMA emphasises likelihood and spreads weight among verbose alternatives. The trade-off is clear: Solomonoff favours simplicity, BMA favours data fit.

\subsection{Per-Cell Probability Distributions}

The per-cell probability grids illustrate how the Solomonoff method and BMA weight their top predictions differently across tasks, particularly when the hypothesis sets are noisy or imperfect.

\subsection{Task A (Alternating Column Shifts)}

In this task, shown in Figure \ref{fig:task-a-weighted-matrix}, the generated hypotheses did not fully capture the true transformation, resulting in a noisy set of competing explanations. As a consequence, the per-cell probabilities reveal more about how each method manages uncertainty than about accurate task modelling. The Solomonoff method assigns moderate confidence values (e.g., 0.65–0.68 in the top-left cells), reflecting its tendency to spread weight across simpler but incomplete hypotheses. BMA, by contrast, often assigns sharper probabilities (closer to 0.9), effectively “collapsing” mass onto whichever noisy hypothesis best fits the data. Because neither method has access to a perfectly faithful hypothesis set, the probability distributions themselves, whether spread by Solomonoff scoring or concentrated by BMA, allow the model to hedge bets across imperfect alternatives, highlighting the value of uncertainty-aware predictions rather than relying on a single, flawed explanation.

\subsection{Task B (Centralisation)}

Here, shown in Figure \ref{fig:task-b-weighted-matrix}, the hypotheses captured the transformation much more faithfully, and both methods achieve near-perfect predictions. The difference lies in calibration: BMA produces very sharp probabilities ($\ge$0.96 in cells containing the central object), while the Solomonoff method leaves more probability mass distributed (e.g., 0.80-0.82 in the same regions). In practice, this means BMA expresses greater certainty, while the Solomonoff simplicity bias results in more conservative assignments. Despite this, both achieve full accuracy, demonstrating that under clean hypothesis sets, either weighting scheme converges on the correct solution.

\subsection{Task C (Alternating column shifts, with 20 competing hypotheses)}

With 20 generated hypotheses, shown in Figure \ref{fig:task-c-weighted-matrix}, this task highlights how each method handles substantial hypothesis noise. Noise refers to the inclusion of multiple competing or imperfect hypotheses, some of which may conflict or only partially explain the data. The Solomonoff method shows greater variability in its probabilities across the grid, with some cells remaining moderately confident (0.49–0.55), indicating unresolved ambiguity. BMA produces sharper outputs (often $\geq$0.8), even in regions where several competing explanations remain plausible. This reflects BMA’s emphasis on likelihood fit, concentrating weight on the best-fitting noisy hypotheses, while the Solomonoff method distributes hypotheses weightings more evenly, leaving visible uncertainty where the set is inconsistent. Both methods performed worse than in the 6-hypothesis setting (Task A), showing that more hypotheses can add unresolved noise. Their failure modes differ: BMA grows overconfident in noisy candidates, while Solomonoff stays underconfident, spreading probability more cautiously. Thus, both degrade with scale, with BMA showing misplaced certainty and Solomonoff showing lingering ambiguity. \

\subsection{Summary}

Across tasks, the trade-off becomes clear: The Solomonoff method is more cautious, often producing better-calibrated but less confident predictions when hypotheses are inconsistent or incomplete. BMA yields sharper, higher-confidence outputs, which can be advantageous when hypotheses are accurate (Task B) but risks overconfidence under noise (Tasks A and C). These results highlight that the true value of both methods lies not in perfect hypotheses, which LLM generation cannot guarantee, but in how probability is distributed across an imperfect set of hypotheses.

\section{Limitations}

\noindent Our approach is limited by the quality of LLM-generated hypotheses. Both Solomonoff-inspired and BMA methods rely on the same candidate pool, which can be noisy, redundant, or partially incorrect. As a result, no single hypothesis perfectly captures the true transformation, necessitating mixtures rather than clear, interpretable rules. Weighting helps filter noise, but performance remains constrained by hypothesis expressiveness and fidelity. \\

\noindent The Solomonoff-inspired method itself also has limitations. Unlike true Solomonoff induction, it uses a finite, biased hypothesis set, substitutes program length with token count, and replaces formal likelihood with cell-wise accuracy. This breaks universality, making results sensitive to tokenisation, prompt design, and evaluation metrics. Computational cost grows with the number of hypotheses, limiting scalability in the Mini-ARC setting. \\

\section{Future Work}
Our results show that Solomonoff-inspired scoring provides calibrated, multi-hypothesis predictions even with noisy candidate sets. Future work can extend these principles beyond Mini-ARC to larger benchmarks such as ARC-AGI-1 and ARC-AGI-2 to test generalisation at scale. This would involve significantly more tasks with statistical testing, sensitivity analyses over hypothesis counts, and ablations to isolate the role of accuracy versus simplicity - studies that are currently computationally prohibitive but are becoming increasingly feasible as more efficient and open-source LLMs emerge. Refinements may also benefit from principles of Minimum Description Length (MDL) \cite{grunwald2007mdl,rissanen1978mdl} and pixel-level uncertainty estimation \cite{martin2023pixelwise}, offering more principled ways to rank hypotheses and calibrate per-cell predictions.

\section{Conclusion}

This work introduced a Solomonoff-inspired framework for ranking and combining LLM-generated hypotheses in Mini-ARC tasks, advancing structured reasoning under uncertainty. By integrating simplicity and predictive accuracy, our approach produced calibrated weightings over noisy candidate sets and, compared to Bayesian Model Averaging, better reflected uncertainty. Solomonoff-inspired scoring offers a principled, practical mechanism for multi-hypothesis reasoning with sparse data, balancing interpretability and effectiveness. While limited by hypothesis quality, our findings highlight the potential of algorithmic information-theoretic priors to guide AI systems toward simpler, more generalisable explanations. This approach can extend to robotic tasks involving uncertainty in perception and action planning, such as selecting grasp trajectories for irregular or cluttered objects in pick-and-place operations or agricultural packing lines. In such settings, multiple feasible strategies often exist, and safety and efficiency depend on reasoning across competing hypotheses. Future work should broaden hypothesis diversity, refine complexity and accuracy measures, and evaluate scalability in richer robotic domains to advance reliable reasoning under uncertainty.

\section*{Acknowledgment}
\noindent The authors would like to thank the Queensland University of Technology (QUT) and the Queensland Centre for Robotics (QCR) for their support in this research. Special thanks to Prof. Will Browne and Cameron Coombe for their supervision and ongoing guidance on my ARC thesis project.

\bibliographystyle{IEEEtran}
\bibliography{references}

\end{document}